\documentclass[a4paper,fleqn]{cas-dc}

\usepackage[authoryear, sort]{natbib}

\def\tsc#1{\csdef{#1}{\textsc{\lowercase{#1}}\xspace}}
\tsc{WGM}
\tsc{QE}

\usepackage{amsmath}
\usepackage{amssymb}
\usepackage{amsfonts}
\usepackage{amsthm}
\usepackage{mathtools}

\usepackage{graphicx}
\usepackage[export]{adjustbox}
\usepackage{wrapfig}
\usepackage{float}
\usepackage{subcaption}

\usepackage{booktabs}
\usepackage{tabularx}
\usepackage{hhline}

\usepackage{bm}
\usepackage{cancel}
\usepackage{oubraces}
\usepackage{optidef}
\usepackage{xfrac}
\usepackage{dsfont}

\usepackage[normalem]{ulem}
\usepackage{paralist}
\usepackage{enumitem}
\usepackage{algorithmic}
\usepackage{textcomp}
\usepackage{comment}
\usepackage{pdfpages}

\usepackage{xcolor}
\usepackage{color}

\usepackage{hyperref}

\DeclareMathOperator*{\argmin}{arg\,min}

\newcommand{\bs}[1]{\boldsymbol{ #1 }}

\theoremstyle{definition}
\newtheorem{definition}{Definition}

\theoremstyle{remark}
\newtheorem{remark}{Remark}

\theoremstyle{definition}
\newtheorem{assumption}{Assumption}
\newtheorem{proposition}{Proposition}

\newcommand{\huseyinrev}[1]{\textcolor[rgb]{0.0, 0.0, 0.0}{#1}}

\newcommand{\huseyin}[1]{\textcolor[rgb]{0.0, 0.0, 0.0}{#1}}
\newcommand{\niko}[1]{\textcolor[rgb]{0.0, 0.0, 0.0}{#1}}

\ExplSyntaxOn

\RenewDocumentCommand \emailauthor { m m }
{
    \int_gincr:N \g_ead_int
    \seq_gput_right:Nn \g_stm_ead_seq
    {
        { \ttfamily \tl_to_str:n {#1} }
    }
}

\ExplSyntaxOff

\begin{document}
\let\WriteBookmarks\relax
\def\floatpagepagefraction{1}
\def\textpagefraction{.001}

\shorttitle{}    

\shortauthors{Dinc et al.}

\title [mode = title]{\textbf{Passivity-Guaranteed Variable Impedance and Compliance Control via Online Parameter Optimization}
}  

\author[1]{Huseyin Tugcan Dinc}[ orcid = 0000-0001-8863-1007 ]
\cormark[1]
\ead{huseyin.dinc@tuwien.ac.at}

\author[2]{Ersin Da\c{s}}[ orcid = 0000-0003-1291-3803 ]
\ead{edas2@illinoistech.edu}

\author[1]{Nikolas Thelenberg}[ orcid = 0009-0005-0323-6275 ]
\ead{nikolas.thelenberg@tuwien.ac.at}

\author[1,3]{Christian Ott}[ orcid = 0000-0003-0987-7493 ]
\ead{christian.ott@tuwien.ac.at}


\affiliation[1]{
    organization={Automation and Control Institute (ACIN), TU Wien},
    city={Vienna},
    postcode={1040},
    country={Austria}
}

\affiliation[2]{
    organization={Department of Mechanical, Materials, and Aerospace Engineering, Illinois Institute of Technology},
    city={Chicago},
    state={IL},
    postcode={60616},
    country={USA}
}

\affiliation[3]{
    organization={Institute of Robotics and Mechatronics, German Aerospace Center (DLR)},
    city={Wessling},
    postcode={82234},
    country={Germany}
}


\cortext[1]{Corresponding author}


\begin{abstract}
Variable impedance control enhances the interaction capability of robot manipulators by adapting impedance parameters online. However, time-varying impedance parameters may inject energy into the system and violate passivity, particularly when tasks demand rapid stiffness changes. Existing methods address this by increasing damping or limiting the stiffness profile, which often degrades the dynamic response. In this work, we propose a quadratic programming-based parameter optimization framework that guarantees passivity by modulating the desired inertia to compensate for energy injected by stiffness variations, while preserving a critically damped system. For the case of diagonal impedance matrices, we optimize the rates of change of inertia and stiffness subject to a passivity constraint formulated as a convex quadratic program. We extend the framework to compliance control, where the desired inertia is coupled to the robot's configuration-dependent inertia matrix. We use an eigendecomposition based on Cholesky factorization to optimize the spectral weights while maintaining passivity. Experiments on the Kinova Gen3 7-degree-of-freedom manipulator validate both methods with human interaction perturbations. Comparisons with an energy tank approach and a nonpassivity-based method demonstrate that the proposed framework achieves less conservative stiffness tracking without requiring an energy reservoir or damping overhead. 
\end{abstract}



\maketitle

\section{Introduction}\label{sec:intro}
Robots operating in real-world environments must physically interact with their surroundings to perform manipulation tasks such as assembly, polishing, or surgical cutting. Variable impedance control (VIC) provides a framework for managing these interactions by allowing robots to continuously modulate the stiffness, damping, and inertia that govern their dynamic relationship with the environment \citep{ ajoudani2016transferring, buchli2011learning}. Through online adaptation of these parameters, robots can transition between compliant behavior when interacting with rigid environments to ensure safety and stiff behavior when manipulating soft objects to maintain precision. In recent years, VIC has attracted growing interest across diverse domains, from human-robot interaction \citep{ficuciello2015variable, hussain2013adaptive} and aerial manipulation \citep{markovic2021adaptive} to whole-body control \citep{dietrich2016whole} and robot learning \citep{abu2020variable, zanella2024learning, abu2018force, rozo2013learning}.

Allowing impedance parameters to evolve online raises a key challenge: guaranteeing closed-loop passivity during parameter adaptation. Unlike fixed-parameter impedance control, where passivity can be ensured with positive definite impedance parameters, VIC requires additional analysis because time-varying impedance modifies the system's energy balance. For instance, abrupt stiffness increases during contact may inject energy and even destabilize the robot--environment interaction if not properly dissipated. Energy tank methods are widely used to enforce passivity constraints in classical impedance control and VIC \citep{schindlbeck2015unified,ferraguti2015energy}. The tank stores dissipated energy and reuses it to compensate for energy injected by stiffness increases \citep{califano2022use}. In \citet{ferraguti2013tank}, an energy tank prevents non-passive stiffness variations for prescribed stiffness and position trajectories. Extensions to redundant manipulators ensured passivity in both task and null spaces \citep{michel2020passivity}, while \citet{michel2023passivity} combined energy tanks with velocity error feedback optimization for stable environmental interaction. In physical human--robot interaction \citep{ferraguti2019variable}, a variable admittance controller was proposed that adapts parameters through an energy tank to suppress oscillations caused by human arm stiffening. Despite their effectiveness, energy tank methods have several drawbacks. The switching logic can introduce jitter, the initial energy budget and tank bounds require careful tuning, and once the tank is depleted, energy injecting impedance adaptations are no longer allowed. 
\begin{figure}  
    \centering
    \includegraphics[width=0.90\columnwidth]{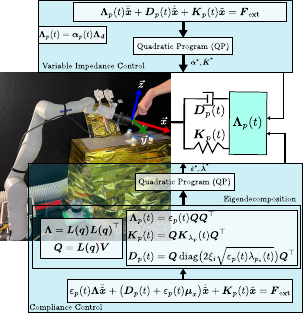} 
    \caption{Overview of the proposed variable impedance and compliance control framework.}
    \label{fig:title_figure}
    \vspace{-4mm}
\end{figure}

Unlike energy tank-based methods, several studies have examined the stability of VIC. In \citet{kronander2016stability}, a Lyapunov-based framework imposed state independent constraints on stiffness and damping profiles. Although stability is guaranteed offline, the method can be conservative for complex stiffness variations. This conservatism was later reduced using passivity filters and modified storage functions with diagonal impedance matrices \citep{bednarczyk2022emg,bednarczyk2020passivity}. Other works addressed exponential stability under bounded disturbances with time-varying impedance \citep{sun2019stability} and established robust stability conditions based on stiffness bounds and rate limits \citep{park2020input}. An MPC framework was proposed that optimizes stiffness eigenvalues and damping ratios to maintain stability during compliance adaptation \citep{thelenberg2024handling}. Another recent work focused on limiting deflection rather than directly constraining stiffness values \citep{muhlbauer2026passive}. A recent virtual inertia-based stabilizing strategy uses the cyclic energy storage behavior of the inertia. The strategy requires switching with strong zero-phase filters to detect sign changes in velocity and acceleration \citep{choi2025unidirectional}. 

We propose variable impedance and compliance control methods for diagonal and \huseyin{general} impedance matrices by optimizing desired inertia to ensure passivity, as illustrated in Fig.~\ref{fig:title_figure}. Our contributions are as follows.\\
(i)~We propose a VIC with time-varying inertia, extending prior constant-inertia formulations and showing that desired-inertia optimization provides an additional mechanism for passivity enforcement.\\
(ii)~We formulate impedance adaptation as an online convex quadratic program (QP) with a state-dependent passivity constraint. Unlike the state-independent conditions in \citet{kronander2016stability}, the proposed method can achieve a wide range of stiffness profiles online without prior impedance information, reducing conservatism and steering the system toward the passive region.\\
(iii)~We extend the optimization-based passivation method to compliance control through an eigendecomposition of the inertia and stiffness matrices. We optimize the inertia scaling factor and reduced system eigenvalues online and reconstruct the desired impedance parameters.\\
(iv)~We experimentally validate the method under large perturbations and significant position errors, demonstrating passive tracking of desired stiffness profiles. Comparison studies show that the proposed method outperforms state-of-the-art passivity-based \citep{ferraguti2013tank} and nonpassivity-based methods \citep{kronander2016stability}.

The remainder of this paper is organized as follows. Section~\ref{sec:pre} presents preliminaries. Section~\ref{sec:var_imp} develops the proposed VIC scheme with passivity guarantees. Section~\ref{sec:compliance} extends VIC to compliance control with optimization-based passivation. Section~\ref{sec:experiments} presents experimental results, while Section~\ref{sec:conc} concludes the work with future directions.
\section{Preliminaries} 
\label{sec:pre}
\subsection{Robot Manipulator Dynamics}
We consider an $n$-degree-of-freedom (DoF), rigid body, \huseyin{redundant} robot manipulator with generalized joint coordinates ${\boldsymbol{q} \in \mathbb{R}^n}$ and joint space dynamics
\begin{equation} \label{joint_space_dynamics}
    \boldsymbol{M}(\boldsymbol{q})\ddot{\boldsymbol{q}} + \boldsymbol{C}(\boldsymbol{q},\dot{\boldsymbol{q}})\dot{\boldsymbol{q}} + \boldsymbol{G}(\boldsymbol{q}) = \boldsymbol{\tau} + \boldsymbol{\tau}_{\mathrm{ext}},
\end{equation}
where ${\boldsymbol{M}(\boldsymbol{q}) \in \mathbb{R}^{n \times n}}$ is the symmetric positive definite inertia matrix, ${\boldsymbol{C}(\boldsymbol{q}, \dot{\boldsymbol{q}}) \in \mathbb{R}^{n \times n}}$ is the Coriolis and centrifugal matrix, and ${\boldsymbol{G}(\boldsymbol{q}) \in \mathbb{R}^n}$ denotes the gravity-induced joint torque vector. The vectors ${\boldsymbol{\tau} \in \mathbb{R}^{n}}$ and ${\boldsymbol{\tau}_{\mathrm{ext}} \in \mathbb{R}^{n}}$ denote the control torques and the external torques acting on the robot, respectively. The matrices~$\dot{\bs{M}}$ and $\boldsymbol{C}$ satisfy the skew-symmetry property such that ${{\bs{v}}^\top({\dot{\boldsymbol{M}}(\boldsymbol{q}, \dot{\bs{q}}) \!-\! 2\boldsymbol{C}(\boldsymbol{q},\dot{\boldsymbol{q}})}){\bs{v}} \!=\! 0}$ for all $\bs{v} \in \mathbb{R}^n$, equivalently,
\begin{equation*} 
    \dot{\boldsymbol{M}}(\boldsymbol{q}, \dot{\bs{q}}) = \boldsymbol{C}(\boldsymbol{q}, \dot{\boldsymbol{q}}) + \boldsymbol{C}(\boldsymbol{q}, \dot{\boldsymbol{q}})^\top.
\end{equation*}
This skew-symmetry property plays a key role in the passivity analysis of the robot dynamics.

\huseyin{Let $\bs{x}=\bs{f}(\bs{q})\in\mathbb{R}^m$, $m\ < n$, denote the end-effector Cartesian coordinates, where $\bs{f}:\mathbb{R}^n\rightarrow\mathbb{R}^m$ is a twice continuously differentiable forward kinematic map. The corresponding velocity and acceleration relationships are
\begin{equation}
\label{eq:task_kinematics}
\begin{aligned}
\dot{\bs{x}}&=\bs{J}(\bs{q})\dot{\bs{q}},\\
\ddot{\bs{x}}&=\bs{J}(\bs{q})\ddot{\bs{q}}
+\dot{\bs{J}}(\bs{q},\dot{\bs{q}})\dot{\bs{q}},
\end{aligned}
\end{equation}
where $\bs{J}(\bs{q})\triangleq\partial\bs{f}/\partial\bs{q}\in\mathbb{R}^{m\times n}$ is the analytic Jacobian.
}
\huseyin{
\begin{assumption}\label{ass:robot_model}
We make the following assumptions throughout the operating region:
\begin{enumerate}
\item The rigid-body dynamic quantities $\boldsymbol{M}(\boldsymbol{q})$, $\boldsymbol{C}(\boldsymbol{q}, \dot{\boldsymbol{q}})$, and $\bs{G}(\bs{q})$ are accurately known.
\item The Jacobian $\bs{J}(\bs{q})$ has full row rank, i.e.,~$\operatorname{rank}(\bs{J}(\bs{q}))=m$.
\item External generalized torques arise from forces at the considered Cartesian interaction port, such that $\bs{\tau}_{\mathrm{ext}}=\bs{J}(\bs{q})^{\top}\bs{F}_{\mathrm{ext}}$, where $\bs{F}_{\mathrm{ext}}\in\mathbb{R}^m$ is known for feedback.
\end{enumerate}
\end{assumption}
}
\huseyin{Under Assumption~\ref{ass:robot_model}}, the Cartesian space inertia matrix and dynamically consistent generalized inverse \citep{khatib1987unified} are defined as
\begin{equation}\label{dyn_consistent_inverse}
\begin{aligned}
\boldsymbol{\Lambda}(\boldsymbol{q}) &\triangleq \bigl(\boldsymbol{J}(\boldsymbol{q})\boldsymbol{M}(\boldsymbol{q})^{-1} \boldsymbol{J}(\boldsymbol{q})^\top\bigr)^{-1} \in \mathbb{R}^{m\times m},\\
\bar{\boldsymbol{J}}(\boldsymbol{q}) &\triangleq \boldsymbol{M}(\boldsymbol{q})^{-1} \boldsymbol{J}(\boldsymbol{q})^\top \boldsymbol{\Lambda}(\boldsymbol{q}) \in \mathbb{R}^{n\times m},
\end{aligned}
\end{equation}
with ${\boldsymbol{J}(\boldsymbol{q}) \bar{\boldsymbol{J}}(\boldsymbol{q}) \!=\! \bs{I}_m}$. 
Due to the redundancy, the acceleration relation~\eqref{eq:task_kinematics} is underdetermined and admits the general solution
\begin{equation} \label{accel_null_space}
    \ddot{\boldsymbol{q}} = \bar{\boldsymbol{J}}(\boldsymbol{q})\bigl(\ddot{\boldsymbol{x}} - \dot{\boldsymbol{J}}(\boldsymbol{q}, \dot{\boldsymbol{q}}) \dot{\boldsymbol{q}}\bigr) + \boldsymbol{N}(\boldsymbol{q}) \ddot{\boldsymbol{q}}_\mathrm{null},
\end{equation}
where ${\ddot{\boldsymbol{q}}_\mathrm{null} \in \mathbb{R}^n}$ is an arbitrary joint acceleration vector, and ${\boldsymbol{N}(\boldsymbol{q}) \triangleq \boldsymbol{I}_n - \bar{\boldsymbol{J}}(\boldsymbol{q}) \boldsymbol{J}(\boldsymbol{q}) \in \mathbb{R}^{n \times n}}$ is the dynamically consistent null space projector. The null space term ${{\boldsymbol{N}(\boldsymbol{q})\ddot{\boldsymbol{q}}_\mathrm{null} }}$ generates internal joint motions that do not affect the instantaneous task space acceleration.

Substituting \eqref{accel_null_space} into \eqref{joint_space_dynamics} and pre-multiplying by $\bar{\boldsymbol{J}}(\boldsymbol{q})^\top$ gives the task space dynamics
\begin{equation} \label{system_model}
\boldsymbol{\Lambda}(\boldsymbol{q}) \ddot{\boldsymbol{x}}
+ \boldsymbol{\mu}_x(\boldsymbol{q},\dot{\boldsymbol{q}})\dot{\boldsymbol{x}} + \boldsymbol{\mu}_v(\boldsymbol{q},\dot{\boldsymbol{q}}){\dot{\bs{q}}} + \bar{\boldsymbol{J}}(\boldsymbol{q})^\top \boldsymbol{G}(\boldsymbol{q}) = \boldsymbol{F} + \boldsymbol{F}_{\mathrm{ext}},
\end{equation}
where
\begin{equation*} 
\begin{aligned}
\boldsymbol{\mu}_x(\boldsymbol{q},\dot{\boldsymbol{q}}) & \!\triangleq\! 
\boldsymbol{\Lambda}(\boldsymbol{q}) \left( \boldsymbol{J}(\boldsymbol{q}) \boldsymbol{M}(\boldsymbol{q})^{-1} \boldsymbol{C}(\boldsymbol{q}, \dot{\boldsymbol{q}}) - \dot{\boldsymbol{J}}(\boldsymbol{q},\dot{\boldsymbol{q}}) \right) \bar{\boldsymbol{J}}(\boldsymbol{q}), \\
\boldsymbol{\mu}_v(\boldsymbol{q},\dot{\boldsymbol{q}}) & \!\triangleq\! 
\boldsymbol{\Lambda}(\boldsymbol{q}) \left( \boldsymbol{J}(\boldsymbol{q}) \boldsymbol{M}(\boldsymbol{q})^{-1} \boldsymbol{C}(\boldsymbol{q}, \dot{\boldsymbol{q}}) - \dot{\boldsymbol{J}}(\boldsymbol{q},\dot{\boldsymbol{q}}) \right) {\boldsymbol{N}}(\boldsymbol{q})
\end{aligned}
\end{equation*}
are the task space Coriolis and centrifugal matrix and the null space coupling matrix, respectively.~The term ${\boldsymbol{\mu}_x(\boldsymbol{q},\dot{\boldsymbol{q}}) \in \mathbb{R}^{m \times m}}$ satisfies ${\dot{\boldsymbol{\Lambda}}(\boldsymbol{q},\dot{\boldsymbol{q}}) = \boldsymbol{\mu}_x(\boldsymbol{q},\dot{\boldsymbol{q}}) + \boldsymbol{\mu}_x(\boldsymbol{q},\dot{\boldsymbol{q}})^\top}$. We define the generalized task space forces as $\boldsymbol{F} \triangleq \bar{\boldsymbol{J}}(\boldsymbol{q})^\top\boldsymbol{\tau} \in \mathbb{R}^{m}$ and ${\boldsymbol{F}_{\mathrm{ext}} \triangleq  \bar{\boldsymbol{J}}(\boldsymbol{q})^\top\boldsymbol{\tau}_{\mathrm{ext}} \in \mathbb{R}^{m}}$.

\subsection{Classical Cartesian Impedance Control}
Given a desired (reference) Cartesian space trajectory ${\bs{x}_d(t) \in \mathbb{R}^{m}}$, 
we define the tracking error as~$\tilde{\bs{x}}(t)  \triangleq \bs{x}(t) - \bs{x}_d(t)$. Then, given time-invariant symmetric positive-definite 
desired inertia, damping, and stiffness matrices denoted by ${\bs{\Lambda}_d \in \mathbb{R}^{m \times m}}$, ${\bs{D}_d \in \mathbb{R}^{m \times m}}$, and
${\bs{K}_d \in \mathbb{R}^{m \times m}}$, respectively,
the dynamic relationship between $\bs{F}_\mathrm{ext}$ and ${\tilde{\bs{x}}}$ is given by
\begin{equation}\label{eq:desired_impedance}
\bs{\Lambda}_d\ddot{\tilde{\bs{x}}}(t) 
+ \bs{D}_d \dot{\tilde{\bs{x}}}(t) 
+ \bs{K}_d \tilde{\bs{x}}(t) = \bs{F}_\mathrm{ext}.
\end{equation}
The task space control force that realizes \eqref{eq:desired_impedance} for the system dynamics in \eqref{system_model} is given by
\begin{equation}
\label{eq:cart_imp_cont}
\begin{aligned}
\bs{F} = {}&\bs{\Lambda}(\bs{q}) \ddot{\bs{x}}_d(t) 
  + \bigl(\bs{\Lambda}(\bs{q}) \bs{\Lambda}_d^{-1} - \bs{I}_m\bigr)\bs{F}_{\mathrm{ext}} \\
  & + \bs{\mu}_x(\bs{q}, \dot{\bs{q}}) \dot{\bs{x}}(t)   
  + \bs{\mu}_v(\bs{q}, \dot{\bs{q}}) \dot{\bs{q}} + \bar{\bs{J}}(\bs{q})^\top \bs{G}(\bs{q}) \\
  & - \bs{\Lambda}(\bs{q}) \bs{\Lambda}_d^{-1}
    \bigl(\bs{K}_d \tilde{\bs{x}}(t) + \bs{D}_d\dot{\tilde{\bs{x}}}(t) \bigr).
\end{aligned}
\end{equation}
\huseyinrev{Feedback of the external force $\bs{F}_\mathrm{ext}$ is essential for inertia shaping \citep{DIETRICH2021104875}.}
The auxiliary null space damping torque is
\begin{equation}\label{eq:null_torque}
\bs{\tau}_\mathrm{null} = -\bs{D}_n \dot{\bs{q}},
\end{equation}
where ${\bs{D}_n \!\in \! \mathbb{R}^{n \times n}}$ is the positive-definite null space damping matrix. The resulting joint torque is
\begin{equation}\label{eq:joint_torque}
\bs{\tau} = \bs{J}(\bs{q})^\top \bs{F} + \bs{N}(\bs{q})^\top\bs{\tau}_\mathrm{null},
\end{equation}
where the projected null space term does not affect the task space dynamics due to the property ${\bar{\bs{J}}(\bs{q})^\top\bs{N}(\bs{q})^\top \equiv \bs{0}}$.
\begin{remark}
\label{re:stable}
In free motion, i.e., when ${\bs{F}_{\mathrm{ext}} = \bs{0}}$, the origin of the closed-loop task space error dynamics, i.e., ${(\tilde{\bs{x}},\dot{\tilde{\bs{x}}})=(\bs{0},\bs{0})}$, in \eqref{eq:desired_impedance} 
is asymptotically stable 
when ${\bs{\Lambda}_d \succ 0}$, ${\bs{D}_d \succ 0}$, and ${\bs{K}_d \succ 0}$. Moreover, the desired impedance dynamics are passive from input $\bs{F}_{\mathrm{ext}}$ to output $\dot{\tilde{\bs{x}}}$ \citep{kronander2016stability}. \huseyinrev{Sections~\ref{sec:var_imp} and~\ref{sec:compliance} analyze passivity of the task-space closed-loop dynamics of redundant manipulators for the input--output pair $(\bs{F}_{\mathrm{ext}},\dot{\tilde{\bs{x}}})$. This analysis does not guarantee passivity of the null space dynamics. See \citet{7368097} for passive null space control.}
\end{remark}

\section{Variable Impedance Control via Impedance Parameter Optimization}
\label{sec:var_imp}
We develop an optimization-based passivation method for VIC with diagonal impedance matrices. 
Section~\ref{sec:compliance} considers compliance control with general impedance matrices. We omit the dependence on configuration variables \huseyinrev{and time} when it is clear from the context. \huseyinrev{Throughout the paper, the subscript~"\textit{d}" denotes desired values, whereas~"\textit{p}" denotes values applied to the controller.} 

\subsection{Variable Impedance Behavior}
VIC specifies a time-varying dynamic relationship between $\tilde{\bs{x}}$ and $\bs{F}_\mathrm{ext}$. We specify the variable Cartesian impedance as
\begin{equation} \label{desired_impedance}
    \boldsymbol{\Lambda}_p(t)\ddot{\tilde{\bs{x}}} + \bs{D}_{p}(t)\dot{\tilde{\bs{x}}} + \bs{K}_{p}(t)\tilde{\bs{x}} = \bs{F}_\mathrm{ext},
\end{equation}
where $\boldsymbol{\Lambda}_p(t)$, $\bs{D}_p(t)$, ${\bs{K}_p(t) \in \mathbb{R}^{m \times m}}$ denote diagonal, time-varying, positive-definite inertia, damping, and stiffness matrices, respectively. 

As discussed in Remark~\ref{re:stable}, the fixed parameter impedance dynamics are asymptotically stable in free motion. However, this argument does not directly extend to VIC since the inertia, damping, and stiffness vary with time. Therefore, the time-varying case requires an additional passivity analysis, which is developed next.

\subsection{Passivity Analysis}
\huseyin{
We first define passivity for the proposed control framework.
\begin{definition}
The time-varying closed-loop system \eqref{desired_impedance} with input $\boldsymbol{F}_\mathrm{ext} \in \mathbb{R}^m$, and output $\dot{\tilde{\bs{x}}} \in \mathbb{R}^m$ is said to be \emph{passive} if, for every admissible input $\boldsymbol{F}_\mathrm{ext}(t)$, the amount of energy that can be extracted from the system over any time interval $[t_0, t_1]$ remains upper-bounded \citep{van2000l2}. 
A continuously differentiable function 
${S_\mathrm{vic} : \mathbb{R}_{\geq 0} \times \mathbb{R}^{m} \times \mathbb{R}^{m} \to \mathbb{R}_{\geq 0}}$ 
with ${S_\mathrm{vic}(t,\bs{0}, \bs{0}) = 0}$ is a storage function if, for all $t_1 \geq t_0 \geq 0$, it satisfies
\begin{equation} \label{passivity_cond}
\begin{aligned}
        S_\mathrm{vic}(t_1, \Tilde{\bs{x}}(t_1), \dot{\Tilde{\bs{x}}}(t_1)) &- S_\mathrm{vic}(t_0, \Tilde{\bs{x}}(t_0), \dot{\Tilde{\bs{x}}}(t_0)) \\
        & \leq \int_{t_0}^{t_1} \dot{\Tilde{\bs{x}}}^\top  \boldsymbol{F}_\mathrm{ext}\;\mathrm{d}t.
\end{aligned}
\end{equation}
\end{definition}
}
\begin{remark} For the input–output pair $(\bs{F}_\mathrm{ext},\dot{\Tilde{\bs{x}}})$ and storage function $S_\mathrm{vic}$, inequality \eqref{passivity_cond} bounds the net increase in stored energy by the energy supplied through the external port. The system can store or dissipate this virtual energy but cannot generate energy internally. 
\end{remark}

We choose the storage function as the sum of kinetic and potential energy:
\begin{equation} \label{sto_func}
    S_\mathrm{vic}(t, \Tilde{\bs{x}}, \dot{\Tilde{\bs{x}}}) = \frac{1}{2}\dot{\Tilde{\mathbf{x}}}^\top\boldsymbol{\Lambda}_p(t) \dot{\Tilde{\bs{x}}} + \frac{1}{2}\Tilde{\bs{x}}^\top\bs{K}_p(t)\Tilde{\bs{x}}.
\end{equation}
We parameterize the inertia matrix $\bs{\Lambda}_p$ as
\begin{equation} 
\label{scaling_inertia}
    \bs{\Lambda}_p(t) = \bs{\alpha}_p(t) \bs{\Lambda}_d,
\end{equation}
where ${\bs{\alpha}_p(t) \succ 0 \in \mathbb{R}^{m \times m}}$ is a diagonal scaling matrix and ${\bs{\Lambda}_d \succ 0 \in \mathbb{R}^{m \times m}}$ is a constant diagonal (user-defined) inertia matrix. The time variation of $\bs{\Lambda}_p(t)$ is determined by $\bs{\alpha}_p(t)$. Taking the time derivative of~\eqref{sto_func} and using~\eqref{desired_impedance} and \eqref{scaling_inertia} yields
\begin{equation} 
\label{S_dot}
\begin{aligned}
    \dot{S}_\mathrm{vic}(t, \Tilde{\mathbf{x}}, \dot{\Tilde{\mathbf{x}}}) &= \frac{1}{2}\dot{\Tilde{\bs{x}}}^\top\dot{\bs{\alpha}}_p (t)\bs{\Lambda}_d\dot{\Tilde{\bs{x}}} + \frac{1}{2}\Tilde{\bs{x}}^\top\dot{\bs{K}}_p(t)\Tilde{\bs{x}}\\
    &- \dot{\Tilde{\bs{x}}}^\top\bs{D}_p(t)\dot{\Tilde{\bs{x}}} + \dot{\Tilde{\bs{x}}}^\top  \bs{F}_\mathrm{ext},
\end{aligned}
\end{equation}
and it has to fulfill~\eqref{passivity_cond} with 
\begin{equation} 
\label{sufficient_cond_passivity}
    \dot{S}_\mathrm{vic}(t, \Tilde{\mathbf{x}}, \dot{\Tilde{\mathbf{x}}}) \leq \dot{\Tilde{\bs{x}}}^\top  \bs{F}_\mathrm{ext}, 
\end{equation}
as it is a sufficient condition for passivity.
Combining~\eqref{S_dot} and~\eqref{sufficient_cond_passivity} leads to the sufficient passivity condition
\begin{equation}
\label{eq:passivity_constraint_vic}
\frac{1}{2}
\dot{\tilde{\bs{x}}}^{\top}
\dot{\bs{\alpha}}_p(t)\bs{\Lambda}_d
\dot{\tilde{\bs{x}}}
+
\frac{1}{2}
\tilde{\bs{x}}^{\top}
\dot{\bs{K}}_p(t)
\tilde{\bs{x}}
-
\dot{\tilde{\bs{x}}}^{\top}
\bs{D}_p(t)
\dot{\tilde{\bs{x}}}
\leq 0.
\end{equation}
Since all matrices are diagonal, \eqref{eq:passivity_constraint_vic} is affine in the diagonal entries of $\dot{\bs{\alpha}}_p(t)$ and $\dot{\bs{K}}_p(t)$, and will be imposed as a constraint in an online optimization problem.

\subsection{Convex Optimization-Based Online Passivation}

At each control instant $t$, let
\huseyin{${\bar{\bs{\alpha}} \triangleq \alpha_p(t)}$ and ${\bar{\bs{K}} \triangleq \bs{K}_p(t)}$} denote the currently applied impedance parameters. \huseyin{Let $\bs{\alpha}_d(t)\succeq \bs{\alpha}_\mathrm{min}$ and $\bs{K}_d(t)\succ0$
denote the user-specified desired parameter trajectories.}
We regulate inertia scaling and stiffness using the first-order reference dynamics 
\huseyin{
\begin{definition}
First-order reference dynamics
\begin{equation} \label{first_order_dynamics}
\begin{aligned}
    \dot{\boldsymbol{\alpha}}_{\mathrm{ref}}\huseyinrev{(t)} &= \dot{\bs{\alpha}}_d\huseyinrev{(t)} + 
         k_{\alpha}(\bs{\alpha}_{d}\huseyinrev{(t)} - \bar{\bs{\alpha}}), \\
    \dot{\bs{K}}_{\mathrm{ref}}\huseyinrev{(t)} &= \dot{\bs{K}}_d\huseyinrev{(t)} + 
         k_{K}(\bs{K}_d\huseyinrev{(t)} - \bar{\bs{K}}),
\end{aligned}
\end{equation}
\end{definition}
}

where $\dot{\boldsymbol{\alpha}}_{\mathrm{ref}}$ and $\dot{\bs{K}}_{\mathrm{ref}}$ are the reference rates of change of inertia scaling and stiffness, and $k_{\alpha}, k_{K} > 0$ are convergence gains. Like~$\bs{\Lambda}_d$, the quantities \(\bs{\alpha}_{d}\), \(\bs{K}_d\), and their rates are desired values provided by the user. Due to the first-order dynamics, the reference rates reduce to the desired rates when tracking errors on inertia scaling and stiffness vanish. Otherwise, the correction terms drive the parameters toward their desired values.

For the optimization, we group decision variables and corresponding reference values in
\begin{equation*} 
    \bs{\Psi}_{v}\huseyinrev{(t)} \!\triangleq\!\! 
    \begin{bmatrix} 
    \operatorname{diag}(\dot{\bs{\alpha}}) \\ 
    \operatorname{diag}(\dot{\bs{K}}) 
    \end{bmatrix} \!\!\in\! \mathbb{R}^{2m},
    \ 
    \bs{\Psi}_{{v}, \mathrm{ref}}\huseyinrev{(t)} \!\triangleq\!\! 
    \begin{bmatrix} 
    \operatorname{diag}(\dot{\bs{\alpha}}_{\mathrm{ref}}) \\ 
    \operatorname{diag}(\dot{\bs{K}}_{\mathrm{ref}}) 
    \end{bmatrix} \!\!\in\! \mathbb{R}^{2m}.
\end{equation*}
\huseyin{The applied parameter rates are obtained by solving the following optimization problem:}
\begin{equation}
\label{quadratic_program_new}
\begin{aligned}
\begin{array}{llll}
\!\!\!\!\!\!\bs{\Psi}^*_v\huseyinrev{(t)} = \displaystyle
    \argmin_{\bs{\Psi}_v} ~~
     (\bs{\Psi}_v - \bs{\Psi}_{v, \mathrm{ref}})^\top\bs{C}_v(\bs{\Psi}_v - \bs{\Psi}_{v, \mathrm{ref}}) \\ [4mm]
~~~~~\textrm{s.t.} 
   ~~ \frac{1}{2}
\dot{\tilde{\bs{x}}}^{\top}
\dot{\bs{\alpha}} \bs{\Lambda}_d
\dot{\tilde{\bs{x}}}
+
\frac{1}{2}
\tilde{\bs{x}}^{\top}
\dot{\bs{K}}
\tilde{\bs{x}}
-
\dot{\tilde{\bs{x}}}^{\top}
\bs{D}_p
\dot{\tilde{\bs{x}}}
\leq 0,
    \\[2mm]
~~~~~~~~~~~\dot{\bs{\alpha}} + \kappa_{\alpha} \bs{\bar{\alpha}}   \succeq \huseyinrev{\kappa_{\alpha} \alpha_\mathrm{min}},
\\[2mm]
~~~~~~~~~~~\dot{\bs{K}} + \kappa_{K} \bs{\bar K}   \succeq 0,
\end{array}
\end{aligned}
\end{equation}
where $\kappa_\alpha ,\kappa_{K} > 0$ bound the admissible decay rates of the impedance parameters, and ${\bs{C}_v \triangleq \operatorname{diag}(\bs{\omega}_\alpha, \bs{\omega}_{K}) \in \mathbb{R}^{2m \times 2m}}$ is the weight matrix that consists of positive weights for each decision variable. \huseyin{The damping matrix is selected as
\begin{equation}\label{eq:critical_damping_vic}
\bs{D}_p =
2\left(
\bar{\bs{K}}\,
\bar{\bs{\alpha}}\,
\bs{\Lambda}_d
\right)^{1/2},
\end{equation}
which yields critical damping along each decoupled
Cartesian axis.
}
\huseyin{The first constraint in~\eqref{quadratic_program_new}
enforces passivity, while the remaining barrier inequalities keep the inertia scaling and stiffness matrices positive semidefinite.} \huseyin{Because the matrices are diagonal, the matrix inequalities are equivalent to elementwise linear inequalities.} We denote the optimal solution of the QP by $${\bs{\Psi}_v^*\huseyinrev{(t)} = \begin{bmatrix} 
    \operatorname{diag}(\dot{\bs{\alpha}}^*) & 
    \operatorname{diag}(\dot{\bs{K}}^*) 
\end{bmatrix}^\top}.$$

We integrate $\bs{\Psi}_v^*\huseyinrev{(t)}$ to obtain the optimal parameters $\bs{\alpha}^*$ and $\bs{K}^*$. We use these values as $\bs{\bar{\alpha}}$ and $\bs{\bar K}$ at the next control instant and repeat the procedure from~\eqref{first_order_dynamics}. We update the impedance parameters as ${\bs{K}_{p} = \bs{K}^*}$, ${\bs{\Lambda}_p = \bs{\alpha}^*\bs{\Lambda}_d}$, and $\bs{D}_p =
2\left(
{\bs{K}^*}\,
\bs{\alpha}^*\,
\bs{\Lambda}_d
\right)^{1/2}$. Finally, we substitute them into the control signal ${\bs{F}_\mathrm{vic}}$:
\begin{equation}
\label{control_law}
\begin{aligned}
\bs{F}_\mathrm{vic} &= \bs{\Lambda}\,\ddot{\bs{x}}_d 
  + \bigl(\bs{\Lambda}\,\bs{\Lambda}_p^{-1} - \bs{I}_m\bigr)\bs{F}_{\mathrm{ext}} 
  + \bs{\mu}_x\,\dot{\bs{x}} 
  + \bs{\mu}_v\,\bs{\dot{q}}  \\
  &\quad + \bar{\bs{J}}^\top\bs{G} - \bs{\Lambda}\,\bs{\Lambda}_p^{-1}
    \bigl(\bs{K}_p\,\tilde{\bs{x}} + \bs{D}_p\dot{\tilde{\bs{x}}}\bigr),
\end{aligned}
\end{equation}
with the null space damping and joint torque composition as 
in~\eqref{eq:null_torque}--\eqref{eq:joint_torque}. Substituting control law (\ref{control_law}) in (\ref{system_model}) yields the \textit{passivated} impedance in (\ref{desired_impedance}).
\huseyin{
\begin{proposition}  \label{prop:qp_feasibility} 
The QP problem~\eqref{quadratic_program_new} together with~\eqref{first_order_dynamics} ensures satisfaction of constraints by optimizing~\(\bs{\Psi}_{v}\huseyinrev{(t)}\). The reference parameter dynamics in~\eqref{first_order_dynamics} ensure exponential parameter convergence when constraints are inactive while the QP in~\eqref{quadratic_program_new} enforces feasible parameter rates to satisfy the constraints. Since the constraints in~\eqref{quadratic_program_new} are linear in the decision variables, and the cost is quadratic, the QP is convex. Therefore, a unique globally optimal solution to the problem exists.
\end{proposition}
\begin{proof}
    Let $\mathcal{F}_v(t)$ denote the feasible set of~\eqref{quadratic_program_new}, expressed as
    \begin{equation*}
    \mathcal{F}_v(t) = \left\{ \bs{\Psi}_v \in \mathbb{R}^{2m} : \bs{\Psi}_v \geq \bs{\ell}_v, \
    \bs{a}_v^\top \bs{\Psi}_v \leq b_v \right\}
    \end{equation*}
    where
    \begin{equation*}
\bs{\ell}_v =
\begin{bmatrix}
- \kappa_\alpha \operatorname{diag} ( \bar{\bs{\alpha}} - \bs{\alpha}_{\min} ) \\
- \kappa_K \operatorname{diag} ( \bar{\bs{K}} )
\end{bmatrix},
    \end{equation*}
\begin{equation*}
\bs{a}_v = \frac{1}{2}
\begin{bmatrix}
\operatorname{diag} ( \bs{\Lambda}_d ) \odot
\dot{\tilde{\bs{x}}} \odot \dot{\tilde{\bs{x}}} \\
\tilde{\bs{x}} \odot \tilde{\bs{x}}
\end{bmatrix},
\quad
b_v = \dot{\tilde{\bs{x}}}^{\top}
\bs{D}_p \dot{\tilde{\bs{x}}},
\end{equation*}
where \(\odot\) denotes element-wise multiplication. Since initially $\bar{\bs{\alpha}} \succeq \bs{\alpha}_\mathrm{min}$ and $\bar{\bs{K}} \succ 0$, we have $\bs{\ell}_v \leq 0$. For $\bs{a}_v \geq 0$ and $b_v \geq 0$, we have $\bs{a}_v^\top \bs{\ell}_v \leq 0 \leq b_v$, implying $\bs{\ell}_v \in \mathcal{F}_v(t)$. The feasible set $\mathcal{F}_v(t)$ is nonempty. It is also closed and convex because all constraints are affine. Since
$\bs{C}_v\succ0$, the objective is continuous,
coercive, and strictly convex. Therefore,
the QP admits a unique global minimizer. 
\end{proof}
}

\section{Optimization-based Compliance Control}
\label{sec:compliance}
In this section, we extend the variable impedance framework of Section~\ref{sec:var_imp} to compliance control. The key distinction is that the applied inertia in the controller is coupled to the robot's actual inertia through a time-varying scaling factor ${\varepsilon_p(t) > 0}$, ${\bs{\Lambda}_p(t) = \varepsilon_p(t)\bs{\Lambda}(\bs{q})}$, preserving the configuration-dependent structure of the original dynamics. \huseyinrev{Since  $\bs{\Lambda}_p$ and $\bs{K}_p\in\mathbb{R}^{m\times m}$ are general symmetric and positive definite matrices}, we parameterize them through \huseyinrev{eigendecomposition} and optimize the spectral weights instead of full matrix entries.

\subsection{\huseyinrev{Eigendecomposition of the Impedance Matrices}} \label{subsec:simul_diag}
In this section, we perform  \huseyinrev{eigendecomposition} of inertia and stiffness matrices similar to \citet{thelenberg2024handling} and \citet{albu2003cartesian}. First, we formulate the compliance behavior as
\begin{equation}\label{compliance_impedance}
\varepsilon_p(t)\bs{\Lambda}\ddot{\tilde{\bs{x}}}
+ \bigl(\bs{D}_p(t) + \varepsilon_p(t)\bs{\mu}_x\bigr)\dot{\tilde{\bs{x}}}
+ \bs{K}_p(t)\tilde{\bs{x}}
= \bs{F}_\mathrm{ext},
\end{equation}
where the term $\varepsilon_p(t)\bs{\mu}_x$ ensures that the skew-symmetry property
\begin{equation}\label{skew_compliance}
{\bs{v}}^\top(\dot{\bs{\Lambda}} - 2\bs{\mu}_x){\bs{v}} = 0 \quad \forall \bs{v} \in \mathbb{R}^m
\end{equation} holds.
Since $\bs{\Lambda}$ is symmetric positive definite, it admits the Cholesky factorization \citep{harville1998matrix}
\begin{equation}
\label{cholesky}
\bs{\Lambda(q)} = \bs{L(q)}\bs{L(q)}^{\top},
\end{equation}
\huseyinrev{where $\bs{L(q)}$ is an invertible lower triangular matrix.} Using this relation, we formulate the eigenvalue problem
\begin{equation}\label{eigenvalue_problem}
\bs{L(q)}^{-1}\bs{K}_p(t)\bs{L(q)}^{-\top}  = \bs{V}\bs{K}_{\lambda_p} \bs{V}^{-1},
\end{equation}
where $\bs{K}_{\lambda_p} = \operatorname{diag}(\lambda_{p_1}, \dots, \lambda_{p_m})$ contains the eigenvalues of $\bs{K}_p$ relative to $\bs{\Lambda}$, and $\bs{V} \in \mathbb{R}^{m \times m}$ is an orthonormal eigenvector matrix ($\bs{V}^{\top}\bs{V} = \bs{I}_m$). Defining 
\begin{equation}
    \bs{Q} \triangleq \bs{L(q)}\bs{V},
\end{equation}
\huseyin{we obtain
\begin{equation}
\bs{\Lambda}=\bs{Q}\bs{Q}^\top, \ \
\bs{K}_p
=\bs{Q}\operatorname{diag}(\bs{\lambda}_p)
\bs{Q}^\top.
\end{equation}
}\huseyin{We parameterize the applied stiffness using the same time-varying basis and optimize its generalized eigenvalues.}
\huseyin{
\begin{assumption} \label{ass:modal_basis}
Stiffness $\bs{K}_p(t)$ and Cartesian inertia $\bs{\Lambda}$ are continuously differentiable and positive-definite over the considered operating interval. The generalized eigenvalues in~\eqref{eigenvalue_problem} are simple and remain separated, and the corresponding eigenvectors are selected continuously such that~$\bs{Q}(t)$ is continuously differentiable. Its derivative $\dot{\bs{Q}}(t)$ is available to the optimization and is independent of the instantaneous optimization variables.
\end{assumption}
}
Under Assumption~\huseyin{~\ref{ass:modal_basis}}, the desired impedance characteristic in~\eqref{compliance_impedance} is reconstructed as  
\begin{equation} \label{new_desired_compliance_impedance}
    \varepsilon_p(t)\bs{Q}\bs{Q}^{\top}\ddot{\tilde{\bs{x}}} + \bigl(\bs{D}_p(t) + \varepsilon_p(t)\bs{\mu}_x\bigr)\dot{\tilde{\bs{x}}} + \bs{Q}\bs{K}_{\lambda_p}(t) \bs{Q}^\top\tilde{\bs{x}} = \bs{F}_\mathrm{ext}.
\end{equation}
In \eqref{new_desired_compliance_impedance}, the impedance parameters appear as
\begin{equation}
\label{decomposed_impedance}
\begin{aligned}
\bs{\Lambda}_p(t) &=  \varepsilon_p(t)\bs{Q}\bs{Q}^{\top},\\
\bs{K}_p(t) &= \bs{Q}\bs{K}_{\lambda_p}(t)\bs{Q}^\top,\\
\bs{D}_p(t) &= \bs{Q}\,\operatorname{diag}\bigl(2\xi_i\sqrt{\varepsilon_p(t)\lambda_{p_i}(t)}\bigr)\bs{Q}^{\top}, 
\end{aligned}
\end{equation}
where $\xi_i > 0$ are modal damping ratios. 
\subsection{Passivity of Compliance Control}\label{sec:Passivity of Compliance Control}
\huseyin{Consider the storage function}
\begin{equation}\label{sto_func_compliance}
S_\mathrm{comp}(t, \Tilde{\mathbf{x}}, \dot{\Tilde{\mathbf{x}}}) = \frac{1}{2}\dot{\tilde{\bs{x}}}^\top\varepsilon_p(t)\bs{\Lambda}\dot{\tilde{\bs{x}}}
     + \frac{1}{2}\tilde{\bs{x}}^\top\bs{K}_p(t)\tilde{\bs{x}}.
\end{equation}
Its time derivative, using the closed-loop dynamics~\eqref{compliance_impedance} and the skew-symmetry property~\eqref{skew_compliance}, is given by
\begin{equation}\label{S_dot_compliance}
\begin{aligned}
    \dot{S}_\mathrm{comp}(t, \Tilde{\mathbf{x}}, \dot{\Tilde{\mathbf{x}}})
&= \frac{1}{2}\dot{\varepsilon}_p(t)\,\dot{\tilde{\bs{x}}}^\top\bs{\Lambda}\dot{\tilde{\bs{x}}}
+ \frac{1}{2}\tilde{\bs{x}}^\top\dot{\bs{K}}_p(t)\tilde{\bs{x}} \\
&- \dot{\tilde{\bs{x}}}^\top\bs{D}_p(t)\dot{\tilde{\bs{x}}} 
+ \dot{\tilde{\bs{x}}}^\top\bs{F}_\mathrm{ext}.
\end{aligned}
\end{equation}
Substituting~\eqref{decomposed_impedance} into~\eqref{S_dot_compliance} yields
\begin{equation} \label{passivity_constraint_with_Qs}
\begin{aligned}
 \dot{S}_\mathrm{comp}(t, \Tilde{\mathbf{x}}, \dot{\Tilde{\mathbf{x}}}) 
 &=  \frac{1}{2}\dot{\tilde{\bs{x}}}^\top\dot{\varepsilon}_p(t)\bs{Q}\bs{Q}^\top\dot{\tilde{\bs{x}}} + \dot{\tilde{\bs{x}}}^\top\bs{F}_\mathrm{ext} \\
 &+ \frac{1}{2}\tilde{\bs{x}}^\top\bs{Q}\operatorname{diag}(\dot{\lambda}_p(t))\bs{Q}^\top\tilde{\bs{x}} + \frac{1}{2}\tilde{\bs{x}}^\top\bs{\Gamma}\tilde{\bs{x}}\\ 
 &- \dot{\tilde{\bs{x}}}^\top\bs{Q}\,\operatorname{diag}\bigl(2\xi_i\sqrt{\varepsilon_p(t)\lambda_{p_i}(t)}\bigr)\bs{Q}^{\top}\dot{\tilde{\bs{x}}}, 
\end{aligned}
\end{equation}
where $\bs{\Gamma} \triangleq \dot{\bs{Q}}\bs{K}_{\lambda_p}\bs{Q}^\top + \bs{Q}\bs{K}_{\lambda_p}\dot{\bs{Q}}^\top$. \niko{The power balance equation~\eqref{passivity_constraint_with_Qs} gives the following sufficient condition for passivity: 
\begin{equation} \label{storage_com_derivative_cond}
    \dot{S}_\mathrm{comp}(t, \Tilde{\mathbf{x}}, \dot{\Tilde{\mathbf{x}}}) \leq  \dot{\tilde{\bs{x}}}^\top\bs{F}_\mathrm{ext}.
\end{equation}
Unlike the diagonal VIC formulation, the passivity condition includes the additional term $\tilde{\bs{x}}^\top\bs{\Gamma}
\tilde{\bs{x}}/2$, which represents the power contribution of eigenbasis variations.}

\niko{
\begin{remark}
    Equation~\eqref{passivity_constraint_with_Qs} contains the eigenvalues~\(\bs{K}_{\lambda_p}\) and eigenvectors~\(\bs{V}\), which we compute numerically. To ensure numerical consistency, consider the permutation matrix~\(\bs{P}\), which sorts the eigenvalues in either ascending or descending order. Then, we obtain the sorted eigenvalues and \mbox{-vectors} via \({\bs{K}_{\lambda_p} \leftarrow \bs{P} \bs{K}_{\lambda_p} \bs{P}^\top}\) and \({\bs{V} \leftarrow \bs{V}\bs{P}^\top}\). To address arbitrary sign changes between consecutive sampling instances, consider the \(i\)\textsuperscript{th}~eigenvector~\(\bs{v}_{i,k-1}\) and~\(\bs{v}_{i,k}\) at the time steps~\(k-1\) and \(k\). If their dot product is negative, i.e. \({\bs{v}_{i,k-1}^\top\bs{v}_{i,k}<0}\), we reverse the sign of the corresponding current eigenvector \(\bs{v}_{i,k}\leftarrow-\bs{v}_{i,k}\).\\
    The aligned eigenvectors are subsequently used to construct $\bs{Q}_k=\bs{L}(\bs{q}_k)\bs{V}_k$, and its time derivative is approximated using the backward difference quotient
    \begin{equation}\label{eq:Q_dot_approx}
        \dot{\bs{Q}}(t_k)\approx
        \frac{\bs{Q}_k-\bs{Q}_{k-1}}{t_k-t_{k-1}}.
    \end{equation}
     This eigenbasis alignment procedure ensures a consistent numerical computation by suppressing permutations and artificial sign discontinuities when evaluating~\(\bs{K}_{\lambda_p}\) and~\(\bs{V}\), maintaining a coherent approximation of~\({\bs{Q}}\) and~\(\dot{\bs{Q}}\).
\end{remark}
}

\subsection{Online Passivation of Compliance Control}
At each control instant, let ${\bar{{\varepsilon}} \triangleq {\varepsilon}(t)}$ and ${\bar{\bs{\lambda}} \triangleq \bs{\lambda}(t)}$ \huseyin{denote currently applied parameters}. The reference rates follow first-order dynamics analogous to~\eqref{first_order_dynamics}:
\begin{equation} \label{first_order_dynamics_compliance}
\begin{aligned}
    \dot{\varepsilon}_{\mathrm{ref}}\huseyinrev{(t)} &= \dot{\varepsilon}_d\huseyinrev{(t)} + k_\varepsilon(\varepsilon_d\huseyinrev{(t)} - \bar{\varepsilon}), \\
    \dot{\bs{\lambda}}_{\mathrm{ref}}\huseyinrev{(t)} &= \bs{\dot{\lambda}}_d\huseyinrev{(t)} + k_\lambda(\bs{\lambda}_d\huseyinrev{(t)} - \bar{\bs{\lambda}}),
\end{aligned}
\end{equation}
where $k_\varepsilon, k_\lambda > 0$ are convergence gains, and $\varepsilon_d$ and $\bs{\lambda}_d$ \huseyinrev{(obtained using~\eqref{eigenvalue_problem})} are desired values. When the tracking errors vanish, the reference rates reduce to the desired rates $\dot{\varepsilon}_d$ and $\dot{\bs{\lambda}}_d$.
As long as the inequality in~\eqref{storage_com_derivative_cond} is satisfied, the closed-loop compliance control system is passive with respect to the input-output pair $(\bs{F}_\mathrm{ext},\dot{\Tilde{\bs{x}}})$.

Similar to $\bs{\Psi}_{v}$ and $\bs{\Psi}_{{v},\mathrm{ref}}$, we define the decision vector and its reference as
\begin{equation*} \label{dec_vars_compliance}
    \bs{\Psi}_{c}\huseyinrev{(t)} \triangleq \begin{bmatrix} \dot{\varepsilon} \\ \dot{\bs{\lambda}} \end{bmatrix} \in \mathbb{R}^{m+1}, \quad
    \bs{\Psi}_{{c},\mathrm{ref}}\huseyinrev{(t)} \triangleq \begin{bmatrix} \dot{\varepsilon}_{\mathrm{ref}} \\ \dot{\bs{\lambda}}_{\mathrm{ref}} \end{bmatrix} \in \mathbb{R}^{m+1},
\end{equation*}
where $\bs{\lambda} \triangleq [\lambda_{1}, \dots, \lambda_{m}]^{\top}$ collects the eigenvalues of $\bs{K}_p(t)$ relative to $\bs{\Lambda}$ as decision variables.
\huseyinrev{Since \eqref{storage_com_derivative_cond} is affine in $\bs{\Psi}_c$}, we construct a convex QP for compliance control:
\begin{equation}
\label{compliance_qp}
\begin{aligned}
\begin{array}{llll}
\!\!\!\!\!\!\bs{\Psi}_c^{*}\huseyinrev{(t)} = \displaystyle
    \argmin_{\bs{\Psi}_c} ~~
    (\bs{\Psi}_c - \bs{\Psi}_{c,\mathrm{ref}})^{\top}
    \bs{C}_c
    (\bs{\Psi}_c - \bs{\Psi}_{c,\mathrm{ref}})
    \\[4mm]
~~~~~~~~~~~\textrm{s.t.}
~~~~~~\dot{S}_\mathrm{comp}(t, \Tilde{\mathbf{x}}, \dot{\Tilde{\mathbf{x}}}) \leq  \dot{\tilde{\bs{x}}}^\top\bs{F}_\mathrm{ext},
    \\[2mm]
~~~~~~~~~~~~~~~~~~~~~
 \dot{{\varepsilon}}
    + \kappa_\varepsilon {\bar{\varepsilon}}
    \succeq \huseyinrev{\kappa_{\varepsilon}} \varepsilon_\mathrm{min},
    \\[2mm]
~~~~~~~~~~~~~~~~~~~~~
   \dot{\bs{\lambda}}
    + \kappa_\lambda \bs{\bar{\lambda}}
    \succeq 0,
\end{array}
\end{aligned}
\end{equation}
where ${\kappa_\varepsilon > 0}$ and ${\kappa_\lambda > 0}$ bound the admissible decay rates of the impedance parameters, $\varepsilon_\mathrm{min}$ is the minimum value of $\varepsilon$, $\bs{C}_c \triangleq \operatorname{diag}(\omega_\varepsilon, \bs{\omega}_\lambda) \in \mathbb{R}^{(m+1) \times (m+1)}$ is the positive-definite weight matrix, and $\bs{\Psi}_c^{*} = \begin{bmatrix} \dot{\varepsilon}^* & \dot{\bs{\lambda}}^{* \top} \end{bmatrix}^\top$ denotes the optimized variables. Integration of $\dot{\varepsilon}^{*}$
and $\dot{\bs{\lambda}}^{*}$ yields $\varepsilon^*$ and $\bs{\lambda}^*$, respectively. We then pass these values to $\bar{{\varepsilon}}$ and $\bar{\bs{\lambda}}$, and repeat the procedure from~\eqref{first_order_dynamics_compliance}.

We compute~$\bs{\Lambda}_p $, $\bs{K}_p$
and $\bs{D}_p$ via~\eqref{decomposed_impedance} using the optimal values $\varepsilon_p = \varepsilon^*$ and $\bs{K}_{\lambda_p} = \operatorname{diag}(\bs{\lambda}^*)$. Combining~\eqref{compliance_impedance} and~\eqref{system_model} gives the control law
\begin{equation}
\label{control_law_compliance}
\begin{aligned}
\bs{F}_\mathrm{comp}
& \!=\! \bs{\Lambda}\,\ddot{\bs{x}}_d
  \!+\! \bs{\mu}_x\,\dot{\bs{x}} 
  \!+\! \bs{\mu}_v\,\bs{\dot{q}}
  \!+\! \bar{\bs{J}}^\top\bs{G}
  \!+\! \left(\frac{1}{\varepsilon_p} \!-\! 1\right)\bs{F}_\mathrm{ext} \\
&\quad - \frac{1}{\varepsilon_p}\bigl((\bs{D}_p + \varepsilon_p(t)\bs{\mu}_x)\dot{\tilde{\bs{x}}}
  + \bs{K}_p\,\tilde{\bs{x}}\bigr),
\end{aligned}
\end{equation}
with the null space damping and joint torque composition as in~\eqref{eq:null_torque}--\eqref{eq:joint_torque}. Substituting control law (\ref{control_law_compliance}) in (\ref{system_model}) yields the \textit{passivated} compliance behavior in (\ref{compliance_impedance}). Note that the constraints in~\eqref{compliance_qp} are linear in the decision variables, making~\eqref{compliance_qp} a QP. Whenever feasible, the strictly convex QP admits a unique global minimizer~\huseyin{as explained below}.

\niko{
\begin{proposition}
Suppose that \(\bar{\varepsilon}\geq\varepsilon_{\min}>0\), \(\bar{\bs{\lambda}}>\bs{0}\), \(\bs{C}_c\succ0\), and
\(\kappa_\lambda =\sup_{\bs{z}\in\mathbb{R}^m\setminus\{0\}} \bs{z}^\top\bs{\Gamma} \bs{z}/\bs{z}^\top \bar{\bs{K}}_p \bs{z}\) hold. Under Assumption~\ref{ass:modal_basis}, the QP~\eqref{compliance_qp} has a globally optimal minimizer and is therefore feasible. 
\end{proposition}
}

\niko{
\begin{proof}
For each time~\(t\), consider the set
\begin{equation}\label{eq:comp_feasible_set}
\mathcal{F}_c(t)= \left\{\bs{\Psi}_c\in\mathbb{R}^{m+1}\;\middle|\;\bs{\Psi}_c\geq\bs{\ell}_c,\, \bs{a}_c^\top\bs{\Psi}_c\leq b_c\right\}
\end{equation}
with the coefficients 
\begin{align}\label{eq:compliance_constraint_coeff}
\begin{split}
\bs{\ell}_c &\triangleq
\begin{bmatrix}
-\kappa_\varepsilon
(\bar{\varepsilon} - \varepsilon_{\min})\\
-\kappa_\lambda\bar{\bs{\lambda}}
\end{bmatrix},\\
\bs{a}_c &\triangleq \frac12
\begin{bmatrix}
\dot{\tilde{\bs{x}}}^{\top}
\bs{\Lambda}\dot{\tilde{\bs{x}}}\\
(\bs{Q}^{\top}\tilde{\bs{x}})
\odot
(\bs{Q}^{\top}\tilde{\bs{x}})
\end{bmatrix},\quad\\
b_c &\triangleq \dot{\tilde{\bs{x}}}^{\top}\bs{D}_p\dot{\tilde{\bs{x}}}-  \frac12\tilde{\bs{x}}^\top \bs{\Gamma}\tilde{\bs{x}},
\end{split}
\end{align}
where \(\odot\) denotes element-wise multiplication. We show that \(\bs{\ell}_c\in\mathcal{F}_c(t)\). First, \(\bs{\ell}_c\geq \bs{\ell}_c\) is always true. The second requirement \({\bs{a}_c^\top\bs{\ell}_c\leq b_c}\) is equivalent to 
\begin{equation} \label{eq:compliance_exact_feasibility}
\tilde{\mathbf{x}}^{\top}   (\bs{\Gamma}-\kappa_\lambda\bar{\bs{K}}_p)\tilde{\mathbf{x}} \leq \dot{\tilde{\mathbf{x}}}^{\top} \left[ 2\bs{D}_p+\kappa_\varepsilon(\bar{\varepsilon}-\varepsilon_{\min})  \bs{\Lambda} \right]\dot{\tilde{\mathbf{x}}}.
\end{equation}
Since the right hand side of the inequality is the sum of two positive-definite matrices, it is positive definite and the expression is always greater than zero. By the assumption on~\(\kappa_\lambda\), we have
\begin{equation}
    \kappa_\lambda =\sup_{\bs{z}\in\mathbb{R}^m\setminus\{0\}} \frac{\bs{z}^\top\bs{\Gamma} \bs{z}}{\bs{z}^\top \bar{\bs{K}}_p \bs{z}}\geq \frac{\tilde{\mathbf{x}}^\top\bs{\Gamma} \tilde{\mathbf{x}}}{\tilde{\mathbf{x}}^\top \bar{\bs{K}}_p \tilde{\mathbf{x}}},
\end{equation}
so the left-hand side of~\eqref{eq:compliance_exact_feasibility} is less than zero. As a result, the inequality~\({\bs{a}_c^\top\bs{\ell}_c\leq b_c}\) is fulfilled and therefore \(\bs{\ell}_c\in\mathcal{F}_c(t)\). The feasible set \(\mathcal{F}_c(t)\) is nonempty. It is also closed and convex, while the objective is continuous, coercive, and strictly convex for \(\bs{C}_c\succ0\). Therefore, the QP admits a unique global minimizer. 
\end{proof}
}

\begin{figure*}
    \centering
    \includegraphics[width=.75\textwidth]{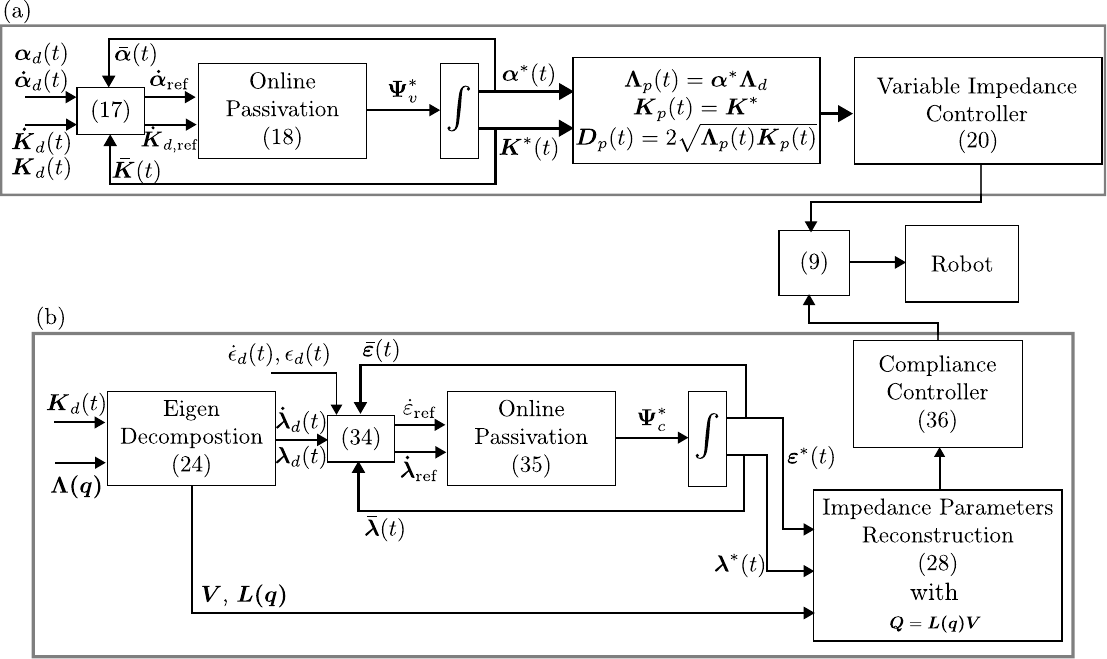}
    \caption{Block diagram representation of the proposed (a) variable impedance controller and (b) compliance controller.}
    \label{fig:block_diagram_final}
\end{figure*}

We optimize the inertia and stiffness parameters within the variable impedance and compliance control framework while preserving critical damping to ensure system performance. Damping can also be included in the optimization problem. We summarize the VIC and compliance controllers in Fig.~\ref{fig:block_diagram_final}. \huseyinrev{At every control cycle, we solve the QPs in Sections~\ref{sec:var_imp} and~\ref{sec:compliance}, whose decision variables are the instantaneous rates of change of the impedance parameters.}

\section{Experimental Validation}
\label{sec:experiments}
We conduct the experiments on a Kinova Gen3 7-DoF manipulator with a Bota SensONE 6-axis force/torque sensor mounted at the end-effector, as shown in Fig.~\ref{fig:title_figure}. The control loop runs at $1$\,[kHz] on an AMD Ryzen 9 9900X processor, and we solve all QP instances in real time using the \textit{proxsuite} library \citep{bambade2022prox}. A model-free disturbance observer from \citet{kim2019model} compensates for unmodeled friction and improves torque tracking. The disturbance observer gains and other experimental parameters are listed in Table~\ref{tab:parameters}. \huseyinrev{We set the stiffness tracking gain higher than the inertia scaling gain so the optimizer preferentially modifies the inertia scaling rate.}
Throughout all experiments, the robot follows a translational sinusoidal Cartesian trajectory
\huseyin{
\begin{equation}
\bs{x}_d = \bs{x}_\mathrm{amp}\sin{(2\pi f t)}    
\end{equation}
where $\bs{x}_\mathrm{amp} \in \mathbb{R}^m$ is the amplitude of the desired position.
}
We validate the proposed framework in three sets of experiments: (i)~variable impedance control with a sinusoidal stiffness profile, (ii)~compliance control with sinusoidal and pulse-wave stiffness profiles, and (iii)~comparison with the energy tank method from \citet{ferraguti2013tank} \huseyinrev{and the Lyapunov-based method from \citet{kronander2016stability}}.

\begin{figure*}
    \centering
    \includegraphics[width=1.0\linewidth]{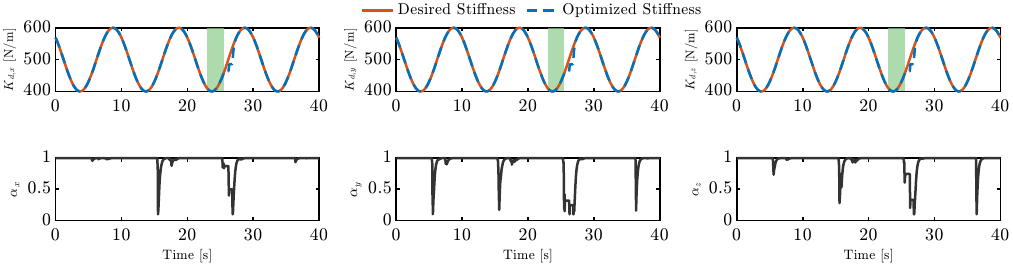}
    \caption{Variable impedance control under a sinusoidal stiffness profile~\eqref{eq:stiffness_profile}. Top row: Optimized (dashed) and desired (solid) stiffness for the $x$, $y$, and $z$ axes; green-shaded regions indicate human interaction at the end-effector. Bottom row: Inertia scaling factors $\alpha_x$, $\alpha_y$, $\alpha_z$: the optimizer reduces $\alpha$ during stiffness increases to maintain passivity.}
    \label{fig:var_imp_subfigure}
\end{figure*}

\begin{figure*}
    \centering
 \includegraphics[width=1.0\linewidth]{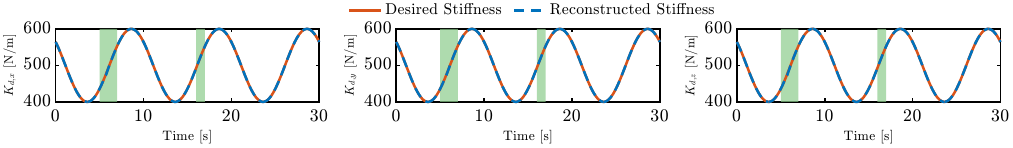}
    \caption{\huseyinrev{Compliance under a sinusoidal stiffness profile. Left to right panels: Desired and reconstructed (actual) stiffness for the $x$, $y$, and $z$ axes via~\eqref{decomposed_impedance}; green-shaded regions indicate human interaction at 5--7\,[s] and 16--17\,[s].}}
    \label{fig:compliance_exp_sine}
\end{figure*}

\begin{figure}
    \centering
    \includegraphics[width=1.0\columnwidth]{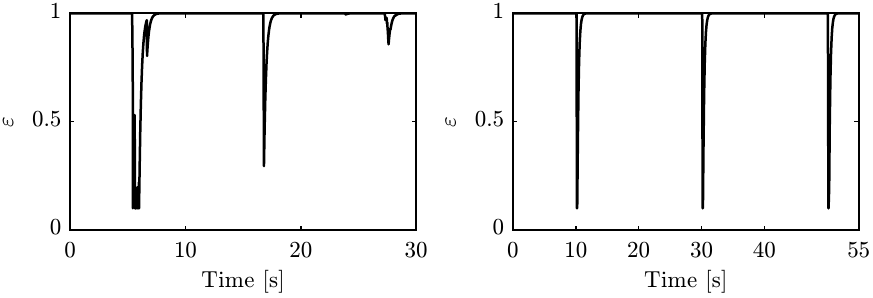}
    \caption{\huseyinrev{Comparison of the inertia scaling factor $\varepsilon$ under different compliance control profiles. Left: Sine-wave profile. Right: Pulse-wave profile.}}
    \label{fig:epsilon_comparison}
\end{figure}

\begin{figure*}
    \centering
    \includegraphics[width=1.0\linewidth]{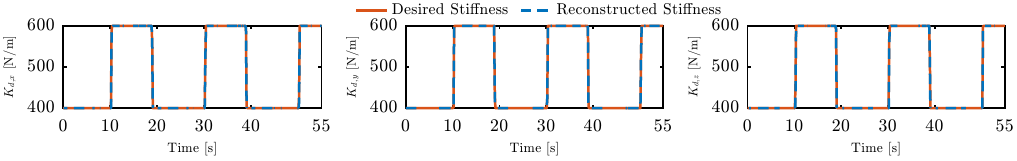}
    \caption{\huseyinrev{Compliance control under a pulse-wave stiffness profile~\eqref{stiffness_pulse} with sharp transitions ($\gamma = 100$). Left to right: Desired and reconstructed stiffness for the $x$, $y$, and $z$ axes. The optimizer tracks the rapid stiffness changes while maintaining passivity through inertia scaling.}}
    \label{fig:compliance_exp_pulse}
\end{figure*}

\subsection{Experiments on Variable Impedance Control}
\label{sec:exp_vic}
The desired stiffness follows the sinusoidal profile
\begin{equation}\label{eq:stiffness_profile}
\bs{K}_d(t) = \bs{K}_0 + \bs{K}_\mathrm{amp}\sin(2\pi f t),
\end{equation}
\huseyin{where $\bs{K}_0=500\bs{I}_3$\,[N/m] and $\bs{K}_{\mathrm{amp}}=100\bs{I}_3$\,[N/m].} The desired stiffness and inertia are set as

$$ 
\begin{aligned}
    \bs{K}_d(t)&= \operatorname{diag}\bigl(\bs{K}_{d,x}(t),\bs{K}_{d,y}(t),\bs{K}_{d,z}(t)\bigr)\ \text{[N/m]}, \\
\quad \bs{\Lambda}_d &=\operatorname{diag}(3,3,3)\ \text{[kg]},
\end{aligned}
$$
along the translational axes.

\begin{figure*}[t]
  \centering
  \includegraphics[width=\textwidth]{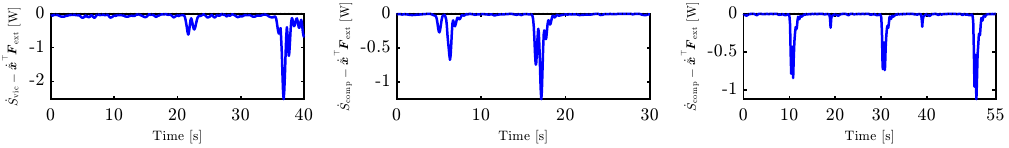}
  \caption{\huseyinrev{Left: Time derivative of the storage function $\dot{S}_\mathrm{vic}$ as in~\eqref{sufficient_cond_passivity} during the variable impedance control experiment. Middle: Compliance control (sinusoidal profile). Time derivative of the storage function $\dot{S}_\mathrm{comp}$ as in~\eqref{storage_com_derivative_cond}. Right: Compliance control (pulse-wave profile). The nonpositive time derivative of the storage functions, confirming passivity throughout.}}
    \label{fig:storage_function_vertical}
\end{figure*}

The top row of Fig.~\ref{fig:var_imp_subfigure} shows the desired and optimized stiffness profiles for the three translational axes. The desired stiffness is tracked throughout the experiment. The bottom row of Fig.~\ref{fig:var_imp_subfigure} shows the corresponding inertia scaling factors $\alpha_i$: during each stiffness increase, the optimizer reduces $\alpha_i$ to compensate for the energy injected into the system, \huseyinrev{implying that the passivity constraint is active}. Between 23 and 25\,[s], a human operator physically interacted with the end-effector, injecting additional energy by increasing the tracking error~$\tilde{\bs{x}}(t)$. \huseyinrev{Since the parameter constraint is also active}, the optimizer responds by reducing the inertia scaling and adjusting the stiffness, maintaining passivity. The damping preserves its critical value $\bs{D}_p = 2(\bs{K}_p\bs{\Lambda}_p)^{1/2}$ throughout the experiment. The passivity constraint~\eqref{passivity_cond} is verified in the first column of Fig.~\ref{fig:storage_function_vertical}, where the time derivative of the storage function remains non-positive for the entire duration. 

\setlength{\tabcolsep}{1pt}
\begin{table}[t]
\centering
\small
\renewcommand{\arraystretch}{0.9}
\caption{Kinova Gen3 7-DoF manipulator experimental demonstration parameters.}
\label{tab:parameters}

\begin{tabularx}{\columnwidth}{@{}l p{0.48\columnwidth} r@{}}
\toprule
\textbf{Symbol} & \textbf{Description} & \textbf{Value} \\
\midrule

\multicolumn{3}{@{}l}{\textit{Common Parameters}}\\
$f_s$ & Control frequency & 1 [kHz] \\
$\bs{L}_{p}$ & Position observer gain & $[75,75,75,75,50,50,50]$ \\
$\bs{L}$ & Velocity observer gain & $[5,5,5,5,4,4,4]$ \\
$f$ & Stiffness and trajectory profile frequency & 0.1 [Hz]\\

\midrule
\multicolumn{3}{@{}l}{\textit{Variable Impedance Control}}\\
$\bs{\alpha}_d$ & Desired inertia scaling & 1.0 \\
$\omega_\alpha$ & Inertia scaling rate weight & 0.5 \\
$\omega_K$ & Stiffness rate weight & 1.0 \\
$\kappa_\alpha$ & $\alpha$ convergence gain & 1.0 \\
$\kappa_K$ & Stiffness convergence gain & 2.0 \\
$K_0$ & Mean stiffness & 500 [N/m] \\
$\bs{K}_\mathrm{amp}$ & Stiffness amplitude & 100 [N/m] \\

\midrule
\multicolumn{3}{@{}l}{\textit{Compliance Control}}\\
$\varepsilon_\mathrm{min}$ & Minimum $\varepsilon$ & 0.10 \\
$\alpha_\mathrm{min}$ & Minimum $\alpha$ & 0.10 \\
$\varepsilon_d$ & Desired inertia scaling & 1.0 \\
$\omega_\varepsilon$ & Inertia scaling rate weight & 0.5 \\
$\omega_\lambda$ & Eigenvalue rate weight & 1.0 \\
$\kappa_\varepsilon$ & $\varepsilon$ convergence gain & 1.0 \\
$\kappa_\lambda$ & Eigenvalue convergence gain & 2.0 \\
$\bs{K}_\mathrm{up}$ & Max. stiffness & 600 [N/m] \\
$\bs{K}_\mathrm{low}$ & Min. stiffness & 400 [N/m] \\
$\gamma$ & Transition sharpness & 100 \\
$\sigma$ & Pulse threshold & 0.2 \\

\bottomrule
\end{tabularx}
\end{table}
\subsection{Experiments on Compliance Control}\label{sec:exp_compliance}
We set the desired stiffness matrix such that 
\huseyin{\begin{equation}\label{eq:desired_nondiagonal_stiffness}
\bs{K}_d(t)=
\begin{bmatrix}
K_{d,x}(t) & 50 & 20\\
50 & K_{d,y}(t) & 20\\
20 & 20 & K_{d,z}(t)
\end{bmatrix}\,\mathrm{[N/m]},
\end{equation}}
\huseyinrev{for the sinusoidal and pulse-wave profiles explained below}, and $\bs{\Lambda}_p = \varepsilon_p(t)\bs{\Lambda}(\bs{q})$. \huseyinrev{The desired eigenvalues are obtained from the desired stiffness matrix $\bs{K}_d(t)$. }\huseyinrev{To keep the eigenvalues~$\bs{K}_{\lambda_p}$ and eigenvectors~$\bs{V}$ consistent, we align the eigenbasis \huseyin{as outlined in Section~\ref{sec:Passivity of Compliance Control}}. This alignment enables us to compute~$\dot{\bs{Q}}$ using a backward difference quotient.} The stiffness matrix is parameterized through its generalized eigenvalues $\lambda_{d_i}$ relative to $\bs{\Lambda}(\bs{q})$. The QP~\eqref{compliance_qp} optimizes $\dot{\varepsilon}$ and $\dot{\bs{\lambda}}$ subject to the passivity \huseyin{and parameter barrier} constraints. The stiffness and damping matrices are reconstructed at each control cycle via~\eqref{decomposed_impedance}. \huseyin{We observed that the term~$\tilde{\bs{x}}^\top\bs{\Gamma}\tilde{\bs{x}}/2$ remained sufficiently small during the experiments. Therefore, we set~$\kappa_\lambda = 2$ throughout.}
\subsubsection{Sinusoidal Profile}
\huseyinrev{The sinusoidal profile~\eqref{eq:stiffness_profile} is used for the diagonal entries of the symmetric positive definite matrix $K_d(t)$ defined above, with $K_0=500\bs{I}_3\ \mathrm{[N/m]}$ and $K_{\mathrm{amp}}=100\bs{I}_3 \ \mathrm{[N/m]}$.}
Figure~\ref{fig:compliance_exp_sine} shows the target and reconstructed stiffness for each axis. The eigenvalue-based parameterization accurately tracks the desired profile. The left panel in Fig.~\ref{fig:epsilon_comparison} shows the inertia scaling factor~$\varepsilon$, which decreases during stiffness increases to maintain passivity. A human operator interacted with the robot at 5--7\,[s] and 16--17\,[s], increasing the tracking errors and injecting additional energy. The optimizer compensates for this energy by reducing $\varepsilon$. The storage function derivative remains non-positive throughout, as shown in the second column of Fig.~\ref{fig:storage_function_vertical}.
\subsubsection{Pulse-Wave Profile}
To test the framework under rapid stiffness changes, we use the pulse-wave profile
\begin{equation} \label{stiffness_pulse}
\begin{aligned}
\bs{K}_d(t) = {}&
\bs{K}_\mathrm{low}
+ \frac{\bs{K}_\mathrm{up} - \bs{K}_\mathrm{low}}{2} \\
& \quad \times
\bigl( \tanh \bigl( \gamma ( \cos ( 2 \pi f t ) - \sigma ) \bigr) + 1 \bigr),
\end{aligned}
\end{equation}
\huseyinrev{which defines the diagonal elements of the stiffness matrix given in Section~\ref{sec:exp_compliance}}, and $\gamma = 100$ produces nearly instantaneous transitions. \huseyinrev{Compared with the sine-wave profile, the pulse-wave profile excites most of the system's modes and provides a more demanding test of the proposed method.} Figure~\ref{fig:compliance_exp_pulse} shows that the optimizer tracks the aggressive stiffness profile. The right panel in Fig.~\ref{fig:epsilon_comparison} shows the inertia scaling $\varepsilon$, which drops sharply during each stiffness step-up to dissipate the injected energy. Passivity is verified in the third column of Fig.~\ref{fig:storage_function_vertical}.
\subsection{\huseyinrev{Comparison with Energy Tanks and Nonpassivity-based Method}}\label{sec:exp_tank}
\begin{figure}
\centering

\begin{subfigure}[t]{0.48\columnwidth}
    \centering
    \includegraphics[width=\linewidth]
        {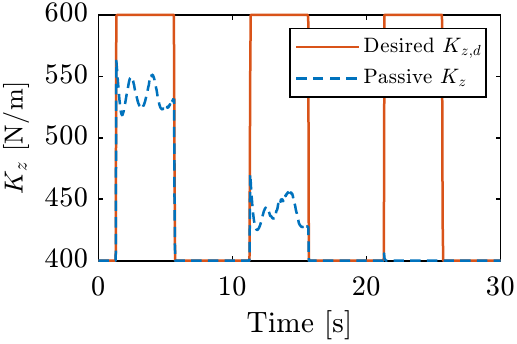}
    \caption{}
    \label{fig:fig1}
\end{subfigure}
\hfill
\begin{subfigure}[t]{0.48\columnwidth}
    \centering
    \includegraphics[width=\linewidth]
    {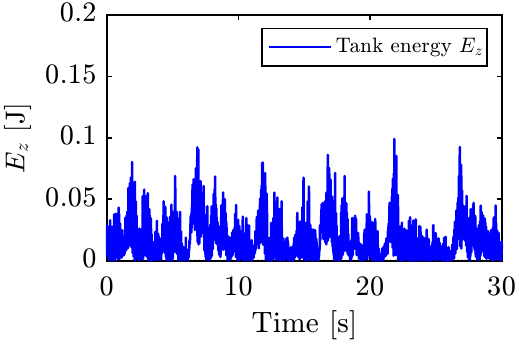}
    \caption{}
    \label{fig:fig2}
            \vspace{-3mm}
\end{subfigure}
\vspace{-1mm}
\caption{Energy tank comparison \citep{ferraguti2013tank} with the pulse-wave stiffness profile. 
(a)~Desired and \huseyinrev{passive} stiffness in the $-z$ axis: once the energy in the tank is insufficient, the controller can no longer track the desired profile and reduces stiffness to maintain passivity. 
(b)~Energy level in the tank, showing depletion during sustained stiffness increases.}
\label{fig:energy_tank}
\vspace{-3mm}
\end{figure}

Using the pulse-wave stiffness profile~\eqref{stiffness_pulse}, we compare the proposed method with the energy tank approach from \citet{ferraguti2013tank}. Fig.~\ref{fig:energy_tank}(a) shows the desired and actual stiffness: once the tank energy is depleted (around 15\,[s]), the controller lacks the stored energy required for the stiffness increase and reduces the stiffness to maintain passivity. The tank energy level, shown in Fig.~\ref{fig:energy_tank}(b), is depleted during the sustained stiffness increases and does not recover sufficiently between pulses. We implement smooth switching using a step-cosine function to avoid the jittery behavior reported in \citet{kronander2016stability} with conventional on-off switching.
\begin{figure}
    \centering
 \includegraphics[width=0.95\linewidth]{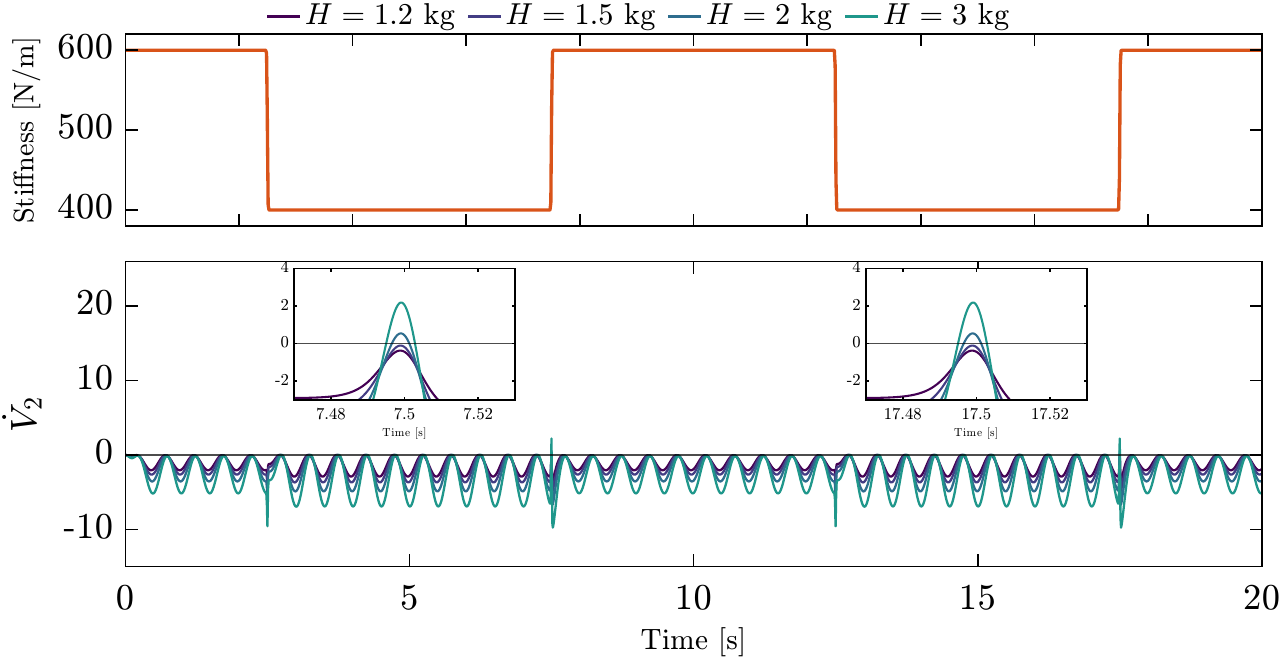}
 \vspace{-1mm}
    \caption{\huseyinrev{Comparison with a nonpassivity-based method \citep{kronander2016stability} using the pulse-wave stiffness profile. The derivative of the Lyapunov function becomes positive when there are sudden switches in the stiffness profile, which violates the stability condition associated with the Lyapunov function.}}
    \label{fig:comparison_with_18}
        \vspace{-3mm}
\end{figure}

In contrast, the proposed QP-based method continuously optimizes the inertia scaling to compensate for the energy injected by stiffness variations, tracking the full stiffness profile without an energy reservoir. The method uses the intrinsic storage function of the impedance dynamics instead of an auxiliary energy budget with finite capacity.

\huseyinrev{We also compared the proposed method with a Lyapunov stability-based technique proposed in \citet{kronander2016stability}.  Following the notation in
that paper, we denote its Lyapunov function and inertia by $V_2$ and $H$, respectively. This method checks offline whether the stiffness profile is achievable while satisfying the stability condition of the designated Lyapunov function. Since this method was designed under a constant inertia assumption, we simulated the scenario and evaluated the derivative of the Lyapunov function for different inertia values while maintaining the damping ratio at $1$ as shown in Fig.~\ref{fig:comparison_with_18}. We observed that when the inertia ${H} > 1.5$ [kg], the derivative of the Lyapunov function $\dot{{V}}_2$ becomes positive, violating the stability condition and preventing the controller from rendering the desired stiffness profile. On the other hand, our proposed method works with a wider range of inertia values and achieves stiffness profiles. }

\section{Conclusion}
\label{sec:conc}
\huseyin{This paper presented an optimization-based framework for passivity-guaranteed variable impedance and compliance control through online modulation of inertia and stiffness parameters. For diagonal impedance matrices, the parameter adaptation was formulated as a convex QP that optimizes the rates of inertia scaling and stiffness, enforces state-dependent passivity and parameter barrier constraints, and preserves critical damping. The feasibility and uniqueness of the optimization problem were established. The framework was further extended to compliance control with configuration-dependent inertia and general stiffness matrices through a generalized eigenvalue decomposition based on Cholesky factorization. The resulting spectral parameterization enables online optimization of inertia scaling and stiffness eigenvalues subject to passivity and parameter barrier constraints, with a single parameter~$\varepsilon$ uniformly scaling inertia.}

\huseyin{Experiments on a Kinova Gen3 7-DoF manipulator demonstrated passive tracking of sinusoidal and pulse-wave stiffness profiles under human interaction perturbations. Compared with energy-tank and Lyapunov-based approaches, the method avoids limitations of energy budgets and switching conditions while accommodating a wide range of stiffness profiles. Future work includes extending the one-step QP formulation to a prediction horizon and investigating orthogonal subspace alignment to improve numerical robustness of the eigenbasis computation in the presence of repeated or closely spaced eigenvalues.}


\section*{CRediT authorship contribution statement}

\noindent
\textbf{Huseyin Tugcan Dinc:}
Conceptualization, Methodology, Software, Formal analysis,
Validation, Writing -- original draft, Visualization.
\textbf{Ersin Da\c{s}:}
Conceptualization, Methodology, Writing -- review \& editing.
\textbf{Nikolas Thelenberg:}
Methodology, Validation, Writing -- review \& editing.
\textbf{Christian Ott:}
Conceptualization, Supervision, Writing -- review \& editing.


\bibliographystyle{cas-model2-names}

\bibliography{cas-refs}


\end{document}